\documentclass{article}

\usepackage{microtype}
\usepackage{graphicx}
\usepackage{subfigure}
\usepackage{booktabs} 
\usepackage{color}
\usepackage{xcolor}

\usepackage{hyperref}

\usepackage[accepted]{icml2021}

\icmltitlerunning{Pretrained Encoders are All You Need}

\begin{document}

\twocolumn[
\icmltitle{Pretrained Encoders are All You Need}



\icmlsetsymbol{equal}{*}

\begin{icmlauthorlist}
\icmlauthor{Mina Khan}{mit}
\icmlauthor{P Srivatsa}{equal}
\icmlauthor{Advait Rane}{bits,equal}
\icmlauthor{Shriram Chenniappa}{bits,equal}
\icmlauthor{Rishabh Anand}{nus}
\icmlauthor{Sherjil Ozair}{deepmind}
\icmlauthor{Pattie Maes}{mit}
\end{icmlauthorlist}

\icmlaffiliation{mit}{MIT Media Lab, Cambridge, MA, USA.}
\icmlaffiliation{bits}{Birla Institute of Technology, Goa, India.}
\icmlaffiliation{nus}{National University of Singapore, Singapore.}
\icmlaffiliation{deepmind}{Deepmind, London, UK.}

\icmlcorrespondingauthor{Mina Khan}{minakhan01@gmail.com}

\icmlkeywords{Unsupervised Representation Learning, Atari, Reinforcement Learning, Spatio-Temporal Attention, Pretrained Models}

\vskip 0.3in
]



\printAffiliationsAndNotice{\icmlEqualContribution} 

\begin{abstract}
Data-efficiency and generalization are key challenges in deep learning and deep reinforcement learning as many models are trained on large-scale, domain-specific, and expensive-to-label datasets. Self-supervised models trained on large-scale uncurated datasets have shown successful transfer to diverse settings. We investigate using pretrained image representations and spatio-temporal attention for state representation learning in Atari. We also explore fine-tuning pretrained representations with self-supervised techniques, i.e., contrastive predictive coding, spatio-temporal contrastive learning, and augmentations. Our results show that pretrained representations are at par with state-of-the-art self-supervised methods trained on domain-specific data. Pretrained representations, thus, yield data and compute-efficient state representations.
\url{https://github.com/PAL-ML/PEARL_v1}
\end{abstract}

\section{Introduction}
Data-efficiency and generalization are key challenges in deep learning (DL) and deep reinforcement learning (RL), especially for real-world deployment and scalability.
Self-supervised learning (SSL) has been used in computer vision and natural language processing \cite{chen_big_2020, radford2021learning, henaff2020data, he2020momentum, devlin_bert_2019, radford2019language} to learn from large-scale unlabeled/uncurated data and do a few-shot transfer to labeled data.
Deep RL has also leveraged self-supervised learning for data-efficiency \cite{srinivas2020curl,laskin2020reinforcement}.

RL in pixel space is also sample-inefficient, and state representations can help sample-efficient, robust, and generalizable RL \cite{lake2016building, kaiser2019model, tassa2018deepmind}.
Previous work investigated self-supervised state representation learning in RL, but involved training on large-scale domain-specific data \cite{anand2019unsupervised}. 

We investigate how pre-trained models can be leveraged for state representation learning in Atari. 
In particular, we focus on using pretrained image representation learning and attention models. 
We also investigate self-supervised fine-tuning of pretrained representations using state-of-the-art (SotA) self-supervised methods.

Our results show that our pretrained representations, particularly `zoomed in' representations, perform as well as the SotA self-supervised state representation learning models trained on large domain-specific datasets. Adding pretrained temporal attention does not particularly improve performance, possibly because the pretrained image representations already leverage attention. Adding self-supervised fine-tuning on domain-specific also does not particularly improve performance, especially in a data-constrained setting.

We make 3 key contributions: 
i. a new methodology for using pretrained image representations for sample-efficient and generalizable state representation learning in RL; 
ii. evaluations of using pretrained temporal attention models with static image representation for temporal data; 
iii. evaluating self-supervised fine-tuning of pretrained representations using SotA SSL techniques on domain-specific data.

\section{Related Work}

Our work lies at the intersection of 5 key areas: Self-Supervised Learning (SSL), Self-Supervised RL, Attention, Domain generalization, and Pretraining. Unlike previous work in Self-Supervised RL, Domain generalization, and Pretraining, we focus on state representation learning in RL, not on the RL. 
Unlike previous work in state representation learning for RL \cite{anand2019unsupervised}, we leverage pretrained models, not trained on domain-specific data, demonstrating data-efficient and generalizable learning. 
Finally, like previous work, we leverage spatial-temporal attention, particularly optical flow \cite{yuezhang2018initial}, and also use SotA SSL, but only for fine-tuning pretrained representations. Our results show that good pretrained embeddings perform competitively, without necessarily needing augmentations from spatio-temporal attention or self-supervised fine-tuning.

\textbf{Self-Supervised Learning (SSL):} 
SSL has been used in natural language processing and computer vision \cite{devlin2018bert, henaff2020data, he2020momentum, chen2020simple} using both contrastive (\cite{oord2018representation, bachman2019learning, he2020momentum, chen_simple_2020, radford2021learning} and purely predictive methods \cite{grill2020bootstrap}. 
SSL on large-scale uncurated datasets has shown promising few-shot transfer to diverse labeled data \cite{chen_big_2020, radford2021learning}. We use pretrained self-supervised models for sample-efficient state representation learning in RL. Also, we use self-supervised fine-tuning of pretrained representations on domain-specific data.

\textbf{Self-Supervised RL:}
SSL has been used in model-based \cite{kaiser2019model, ha2018world, schrittwieser2020mastering, hafner2019dream} and model-free RL \cite{jaderberg2016reinforcement, shelhamer2016loss,oord2018representation}. 
Learning is either reconstruction-based  \cite{jaderberg2016reinforcement, higgins2017darla, yarats2019improving} or constrastive \cite{sermanet2018time, warde2018unsupervised, anand2019unsupervised, oord2018representation, srinivas2020curl}, and the predictions are in pixel space \cite{jaderberg2016reinforcement} or latent space \cite{oord2018representation, guo2020bootstrap}. Image augmentations  \cite{srinivas2020curl,laskin2020reinforcement}, spatio-temporal structures \cite{sermanet2018time, aytar2018playing, oord2018representation, anand2019unsupervised}, and scene or object representations \cite{burgess2019monet, zhu2018object, greff2019multi, van2019perspective} have also been used.
Representation learning has also been decoupled from RL \cite{anand2019unsupervised, stooke2020decoupling}. We focus on state representation learning in RL, instead of the RL itself, and unlike previous work, leverage pretrained self-supervised representations not trained on domain-specific data.

\textbf{Attention:} Attention has been used in RL \cite{zhang2018agil, gregor2018visual, manchin2019reinforcement, salter2019attention}, especially to add robustness and interpretability \cite{sorokin2015deep,mott2019towards}. Optical flow-based attention has also been used in RL \cite{yuezhang2018initial}. We explore spatio-temporal attention, including optical flow, to improve image representations for temporal data.

\textbf{Domain generalization:} Transfer learning uses representations from one domain to generalize to different domains \cite{rusu2016sim, oquab2014learning, donahue2013decaf}. Domain adaptation adapts from one domain to another domain using data from the target domain \cite{bousmalis2016domain,ganin2016domain, wulfmeier2017mutual}, whereas
domain randomisation covers a distribution of environments during training to generalize \cite{sadeghi2016cad2rl, andrychowicz2020learning, viereck2017learning, held2017probabilistically, tobin2017domain, peng2018sim}. We use pretrained models, not trained on Atari-specific data, for sample-efficient state representation learning in RL.

\textbf{Pretraining:} Unsupervised pretraining has been leveraged in RL, e.g., by maximizing diversity of states \cite{liu2021behavior} or skills \cite{eysenbach2018diversity,hansen2019fast,sharma2019dynamics}. Intrinsic rewards for exploration \cite{pathak2016context, sekar2020planning}, predicting state dynamics \cite{anderson2015faster}, and reward-free representations \cite{schwarzerpretraining} have also been used. Transferable skills can also be learned \cite{campos2021coverage}. We use pretrained models for state representation learning in RL.

\section{Approach}

\textbf{State representation learning in RL:} Deep reinforcement learning has leveraged the expressive power of deep learning to create end-to-end models for RL in high dimensional spaces.
RL in high dimensional spaces, e.g., pixels spaces, however, is sample-inefficient \cite{lake2016building, kaiser2019model} and learning policies from state representations could be more \textbf{sample-efficient, robust, and generalizable} \cite{tassa2018deepmind, liu2021return, eslami2018neural}.
Thus, we decouple state representation learning from RL \cite{stooke2020decoupling} and aimed to investigate state representation learning in RL \cite{anand2019unsupervised}. 

\textbf{Pretrained models with self-supervised fine-tuning:} 
Pretrained models have been leveraged for few-shot learning \cite{chen_big_2020, henaff2020data, radford2021learning}, but not for state representation learning in RL.
State representation learning in RL has leveraged SSL \cite{anand2019unsupervised}, but involved domain-specific and data-intensive training.
We aimed to leverage pretrained models for sample-efficient state representation learning in RL. We decided on 3 key explorations:
\textit{i. Pretrained image representations}, including `zooming in' via grid-based patches;
\textit{ii. Spatio-temporal attention}, e.g., using optical flow;
\textit{iii. Self-supervised fine-tuning} on domain-specific data via constrastive losses.

\begin{figure}
    \centering
    \includegraphics[width=\linewidth]{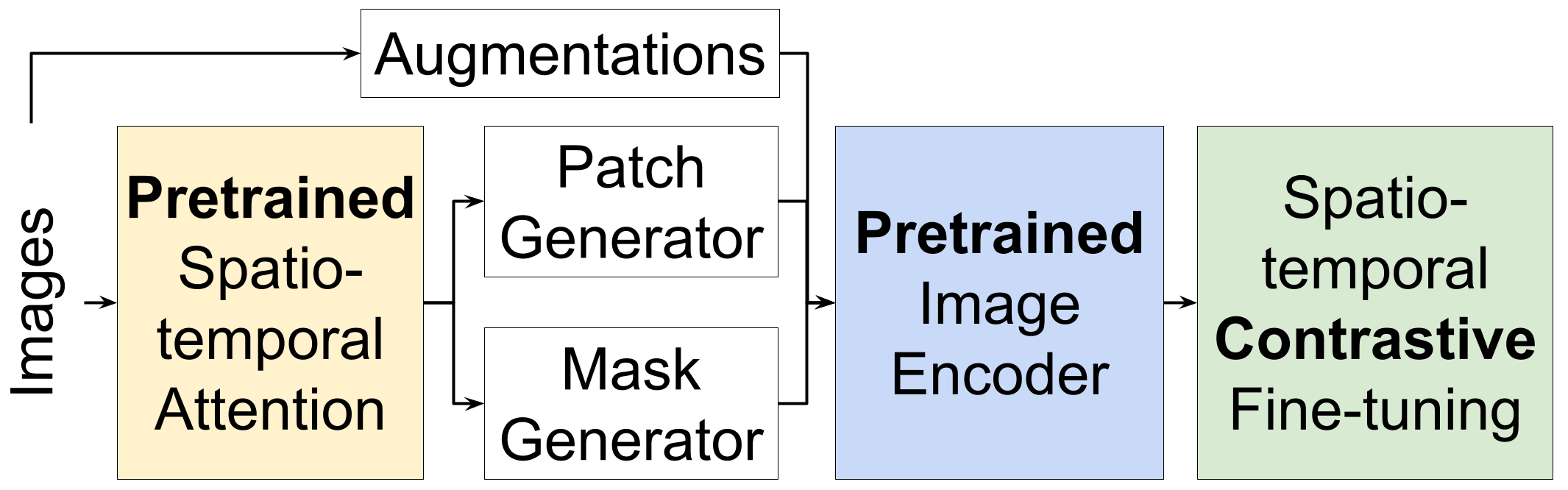}
    \caption{PEARL: Pretrained Encoder \& Attention for Representation Learning.}
    \label{fig:architecture}
\end{figure}

Our framework, \textbf{PEARL} (Pretrained Encoder and Attention for Representation Learning), has 3 components (Figure \ref{fig:architecture}).

\subsection{Pretrained Image Representations} 
Recent work compared image representations from different supervised, self-supervised, and weakly supervised models, and found that CLIP \cite{radford2021learning}, a weakly supervised image representation learning model, produced the best representations for few-shot learning \cite{khan2021personalizing}. Thus, we selected CLIP as our image encoder.

In addition to image representations for full-image, we also considered `zooming in' via grid-based patches, i.e., equally-sized non-overlapping patches covering the whole image.

\subsection{Spatio-temporal Attention}
For spatial attention, we considered a SotA supervised object detection model, i.e., EfficientDet \cite{tan2019efficientdet}, and a SotA SSL model, i.e., Dino \cite{caron2021emerging}.

For temporal attention, we decided to compare optical flow using RAFT, a SotA pretrained model \cite{teed2020raft}, with image difference using structural similarity \cite{wang2004image}. We used image difference particularly because objects appear/disappear in video animations and optical flow may not be able to capture sudden flowless changes. 

Finally, we compare two types of attention: i. mask-based attention, which highlights the relevant regions in the full image; ii. patch-based attention, which crops out the relevant patches and calculates embeddings for each by zooming in.

\subsection{Self-supervised Fine-tuning} We considered 3 SotA methods for self-supervised fine-tuning of pretrained representations using domain-specific data: i. Image augmentations (color jitter, random crop, and gaussian blur) \cite{chen_simple_2020}; ii. Spatio-temporal constrastive learning (ST-DIM) \cite{anand2019unsupervised}; iii. Contrastive predictive learning (CPC) \cite{oord2018representation}.

\section{Experiments}

We outline our experiments and findings below.
Similar to the SotA work \cite{anand2019unsupervised}, our evaluations use a linear probe \cite{alain_understanding_2018} with a \{70, 10, 20\}\% \{train, validation, and test\}-split with 50k data points and early stopping. We share the F1 score for each Atari game.

\begin{figure}
    \centering
    \includegraphics[width=\linewidth]{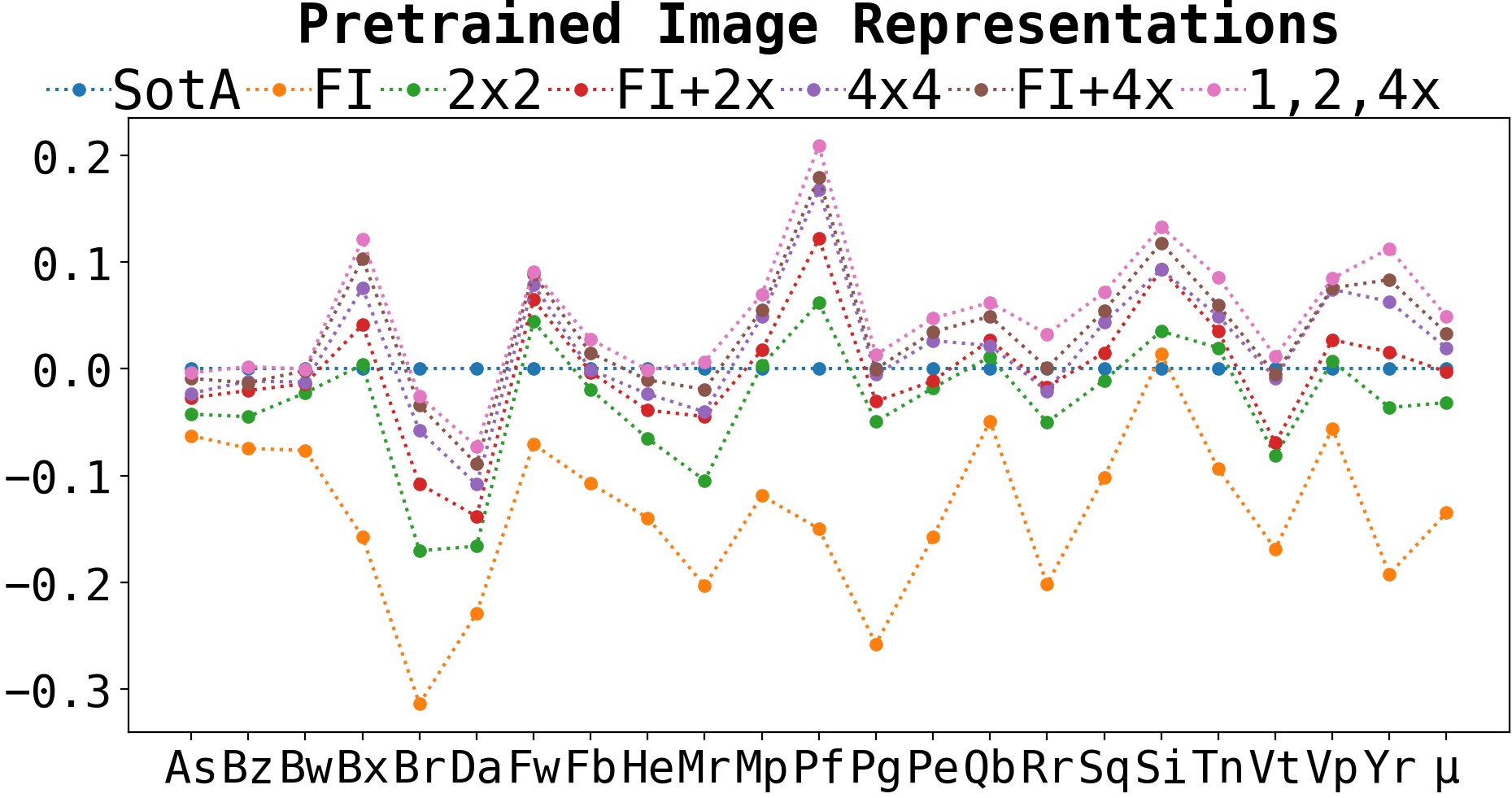}
    \caption{SotA vs pretrained representations for patches (vanilla).}
    \label{fig:patches}
\end{figure}

\subsection{Pretrained Image Representations} \label{vanilla}
\textit{Setup:} We tried 3 nxn-grid patch sizes (1x1, 2x2, and 4x4). Each patch generated a 512-size embedding using CLIP and we concatenated the embeddings. We tried 6 different configurations: 1. 1x1, i.e., full image (FI); 2. 2x2; 3. 1x1+2x2; 4. 4x4; 5. 1x1+4x4; 6. 1x1+2x2+4x4. Figure \ref{fig:patches} compares the results for our 6 configurations to the SotA results.
SotA refers to the top performance of all the models in \cite{anand2019unsupervised} -- ST-DIM was top in all, except 4 games, CPC was top in 3, and Pixel-Pred was top in 4.
 
\textit{Results:} On average, using at least 5 embeddings from 1 full image and 4 2x2 patches, CLIP's pretrained embeddings perform better than SotA. Even our worst configuration (1x1) is on average better than a VAE trained on domain-specific data \cite{anand2019unsupervised}. With 21 patches (1x+2x+4x), our model performs better than the SotA in all, except 2 games, and has an average of 5\% better performance than SotA. Overall, the performance improves with increasing number of patches. Thus, zooming in gives performance improvements at the cost of bigger embedding sizes.

\begin{figure}
    \centering
    \includegraphics[width=\linewidth]{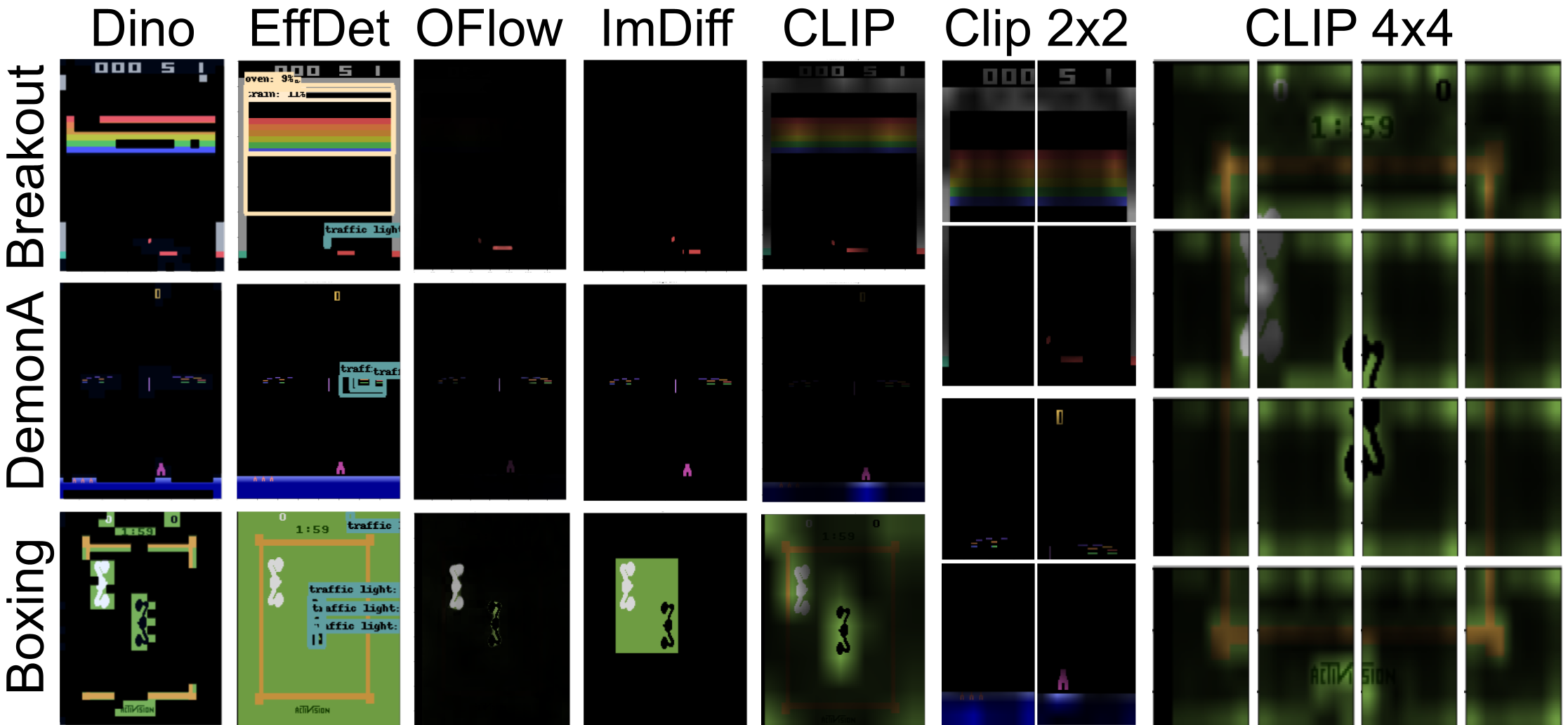}
    \caption{Spatio-temporal attention masks from pretrained models.}
    \label{fig:attn_viz}
\end{figure}

\subsection{Spatial-Temporal Attention} 

\textit{Setup:} We compared Optical Flow Mask (FM) with Image Difference Mask (DM), and also, each mask combined with full image, i.e., FM+ (FM+FI) and DM+ (DM+FI). 
We evaluated patch-wise attention using full image combined with 4 patches selected from 4x4 and 2x2 patches, weighted by Optical Flow (FP5) and Image Difference (DP5).
Each image `combination' refers to a concatenation of the image embeddings. 
We compared each of our 3 settings with their equivalent size vanilla embeddings, i.e., embeddings using just the pretrained encoder (Section \ref{vanilla}): i. DM and FM with FI (1x512 embedding); ii. DM+ and FM+ with FI+FI (2x512 embedding); iii. DP5 and FP5 with FI+2x2 (5x512 embedding). We did not evaluate EfficientDet and Dino as the attention masks did not look promising (Figure \ref{fig:attn_viz}).

\begin{figure}
    \centering
    \includegraphics[width=\linewidth]{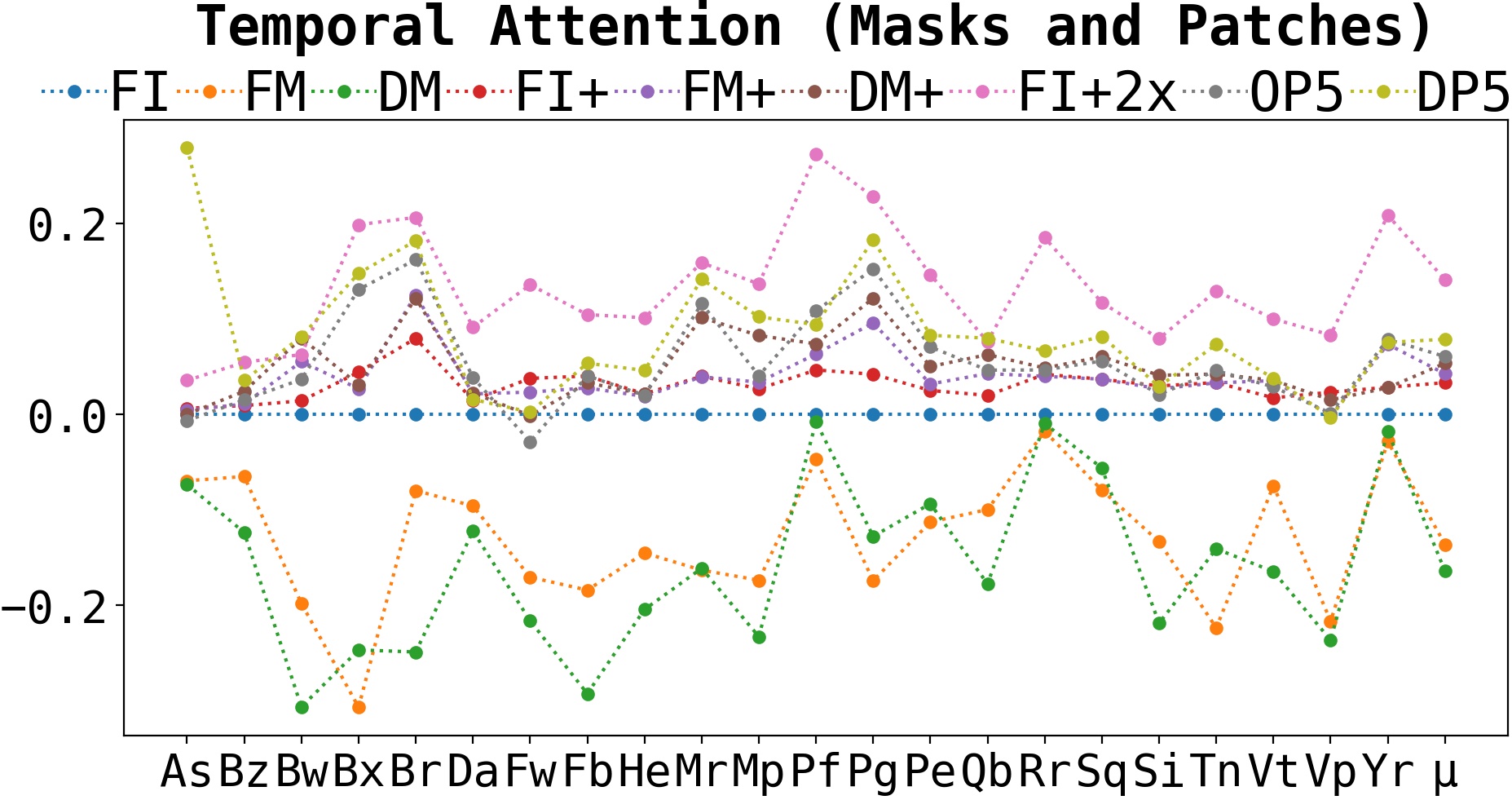}
    \caption{Temporal attention masks and patches vs vanilla results.}
    \label{fig:t_attn}
\end{figure}

\textit{Results:} Our results (Figure \ref{fig:t_attn}) show that optical flow masks (FM) alone are on average better than image difference (DM) masks. However, image difference masks combined with the full image (DM+) are better than optical flow masks combined with the full image (FM+). Also, both FM+ and DM+ are both slightly better than the two full image embeddings concatenated (FI+), and the difference is biggest in games like Pong and Montezuma's Revenge. Finally, full image + 4 non-grid patches from image difference (DP5) are slightly better than full image + 4 non-grid patches from optical flow (FP5), but they are both worse than the full image + 4 2x2 grid patches. 
Thus, n grid-based patches may be better than n non-grid patches, which may zoom in more but not cover the full image. Also, temporal masks combined with full image are slightly better than two copies of the full image. However, the difference is not significant and is game-dependent, possibly because our encoder model, CLIP, already incorporates `good enough' attention (Fig \ref{fig:attn_viz}). 

\subsection{Self-supervised Fine-tuning}

We evaluated if state-of-the-art self-supervised methods could be used to fine-tune pretrained embeddings using domain-specific data. We evaluated three self-supervised methods: i. Augmentations (random crop, color jitter, and gaussian blur) with an MLP layer and pairwise contrastive loss \cite{chen_big_2020}; ii. ST-DIM \cite{anand2019unsupervised} with separate temporal (T-DIM), spatial (S-DIM), and spatio-temporal contrast (ST-DIM); iii. CPC, replacing CPC's original encoder with our CLIP encoder and then adding a linear layer followed by a Gated Recurrent Unit. We compare each of the evaluations with their equivalent embedding-size baselines from Section \ref{vanilla} -- FI for Blur, Jit, Crop, CPC, and T-DIM-1x (T1x); 2x2 for T-DIM 2x2 (T2x) and S-DIM (s2x); 1x1+2x2 for ST-DIM 1x1+2x2 (ST1,2x).

\begin{figure}
    \centering
    \includegraphics[width=\linewidth]{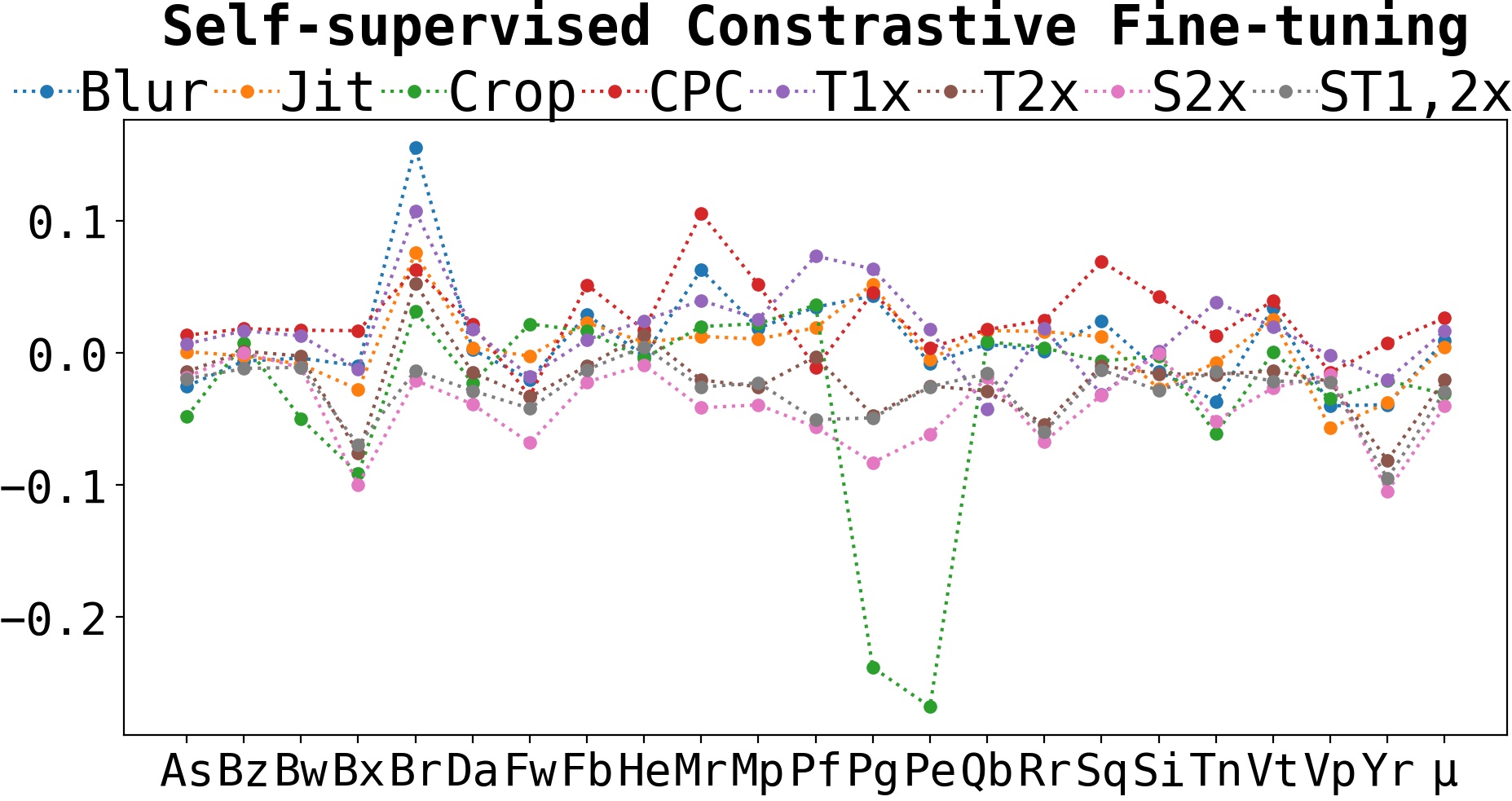}
    \caption{Performance changes using self-supervised fine-tuning}
    \label{fig:constrastive}
\end{figure}

Our results (Figure \ref{fig:constrastive}) show that on average, fine-tuning with CPC leads to the most improvement (3\%), followed by T-DIM-1x1 and gaussian blur. Color jitter, on average, makes no change and the rest lead to a 2-4\% drop in performance compared to the equivalent vanilla cases (Section \ref{vanilla}). Maximum improvements are using gaussian blur in Breakout (15.5\%), CPC in Montezuma's revenge (10.8\%), and T-DIM in Breakout (6.3\%), but most improvements are game and method-dependent. Thus, overall, there are no significant improvements using self-supervised fine-tuning, possibly due to data or model constraints and especially since we are not fine-tuning the pretrained CLIP model. 

\section{Conclusion}

Data-efficiency and generalization are key challenges in deep learning and deep RL. 
Most deep RL models are trained from scratch on domain-specific data, but recent research shows that self-supervised models trained on large-scale uncurated data have promising few-shot transfer.

We investigated the use of pretrained models for state representation learning in RL.
Our results show that pretrained self-supervised models, not trained on domain-specific data, give competitive performance compared to SotA self-supervised models, trained on large-scale domain-specific data. Thus, pretrained models enable data-efficient and generalizable state representation learning for RL. 

Moreover, our framework, PEARL (Pretrained Encoder and Attention for Representation Learning), investigates not only using pretrained image representations but also pretrained spatio-temporal attention. Our pretrained image representation model also uses attention and our results show that attention helps state representation learning. 

Finally, even though self-supervised fine-tuning on domain-specific data did not significantly improve pretrained representations, it could be because we froze our pretrained model and had a few trainable layers with limited data. 

State representations are key in RL. We believe that `Pretrained Encoders are All You Need' (PEAYN) for data-efficient and generalizable state representation learning in RL, and will hopefully be a stepping stone to data-efficient, generalizable, and interpretable RL. No PEAYN, no gain :)

\bibliography{example_paper}
\bibliographystyle{icml2021}

\pagebreak
\appendix

\section{Atari Game Abbreviations}
Table \ref{tab:games} shows the list of all games and their respective abbreviations used in our paper.
\begin{table}[]
    \centering
    \begin{tabular}{ll}
    \toprule
    Abbreviation & Game Name \\
    \bottomrule
         As & Asteroids \\
    Bz & Berzerk \\
    Bw & Bowling \\
    Bx & Boxing \\
    Br & Breakout \\
    Da & DemonAttack \\
    Fw & Freeway \\
    Fb & Frostbite \\
    He & Hero \\ 
    Mr & Montezuma’s Revenge \\
    Mp & MsPacman \\
    Pf & Pitfall \\
    Pg & Pong \\
    Pe & PrivateEye \\
    Qb & Qbert \\ 
    Rr & Riverraid \\
    Sq & Seaquest \\
    Si & SpaceInvaders \\
    Tn & Tennis \\ 
    Vt & Venture \\
    Vp & VideoPinball \\
    Yr & YarsRevenge \\ 
    $\mu$ & Average \\
    \bottomrule
    \end{tabular}
    \caption{Caption}
    \label{tab:games}
\end{table}


\end{document}